\documentclass{article}

\usepackage{arxiv}

\usepackage[utf8]{inputenc} 
\usepackage[T1]{fontenc}    
\usepackage{hyperref}       
\usepackage{url}            
\usepackage{booktabs}       
\usepackage{amsfonts}       
\usepackage{nicefrac}       
\usepackage{microtype}      
\usepackage{lipsum}
\usepackage{graphicx}
\graphicspath{ {./images/} }
\usepackage{subcaption}
\usepackage[ruled,vlined]{algorithm2e}
\usepackage{optidef}

\title{Multi-View Matching (MVM): Facilitating Multi-Person 3D 
Pose Estimation Learning with Action-Frozen People Video}

\author{
 Yeji Shen \\
  University of Southern California \\ Los Angeles, CA 90089, USA \\
  \texttt{yejishen@usc.edu} \\
   \And
 C.-C. Jay Kuo \\
  University of Southern California \\ Los Angeles, CA 90089, USA \\
  \texttt{cckuo@sipi.usc.edu} \\
}

\begin{document}
\maketitle
\begin{abstract}
To tackle the challeging problem of multi-person 3D pose
estimation from a single image, we propose a multi-view matching (MVM)
method in this work. The MVM method generates reliable 3D human poses
from a large-scale video dataset, called the Mannequin dataset, that
contains action-frozen people immitating mannequins.  With a large
amount of in-the-wild video data labeled by 3D supervisions
automatically generated by MVM, we are able to train a neural network
that takes a single image as the input for multi-person 3D pose
estimation.  The core technology of MVM lies in effective alignment of
2D poses obtained from multiple views of a static scene that has a
strong geometric constraint. Our objective is to maximize mutual
consistency of 2D poses estimated in multiple frames, where geometric
constraints as well as appearance similarities are taken into account
simultaneously.  To demonstrate the effectiveness of 3D supervisions
provided by the MVM method, we conduct experiments on the 3DPW and the
MSCOCO datasets and show that our proposed solution offers the
state-of-the-art performance. 

\end{abstract}


\section{Introduction}\label{sec:introduction}

Human pose estimation is a long-standing problem in computer vision
research. It has numerous applications such as sports, augmented
reality, motion analysis, visual avatar creation, etc. In the past ten
years, a major advance has been made in building large-scale human pose
datasets and developing deep-learning-based models for human pose
estimation. As a result, estimating multi-person 2D poses and/or a
single-person 3D pose in complicated scenes become mature. Yet,
multi-person 3D pose estimation is still a challeging problem due to the
lack of large-scale high-quality datasets for this application. This is
further hindered by inadequate depth estimation algorithms in monocular
images. 

It is nontrivial to acquire high quality 3D supervisions in dynamic
scenes. Depth sensors (e.g., Kinect) can provide useful data, yet the
acquisition is typically limited to indoor environments. Furthermore, it
often demands a large amount of manual work in capturing and processing.
As an alternative, Li {\em et al.} \cite{li2019learning} attempted to
estimate dense depth maps using the multi-view stereo (MVS) method from
video clips captured in stationary scenes. 

\begin{figure}[h]
\centering
\includegraphics[width=0.72\textwidth]{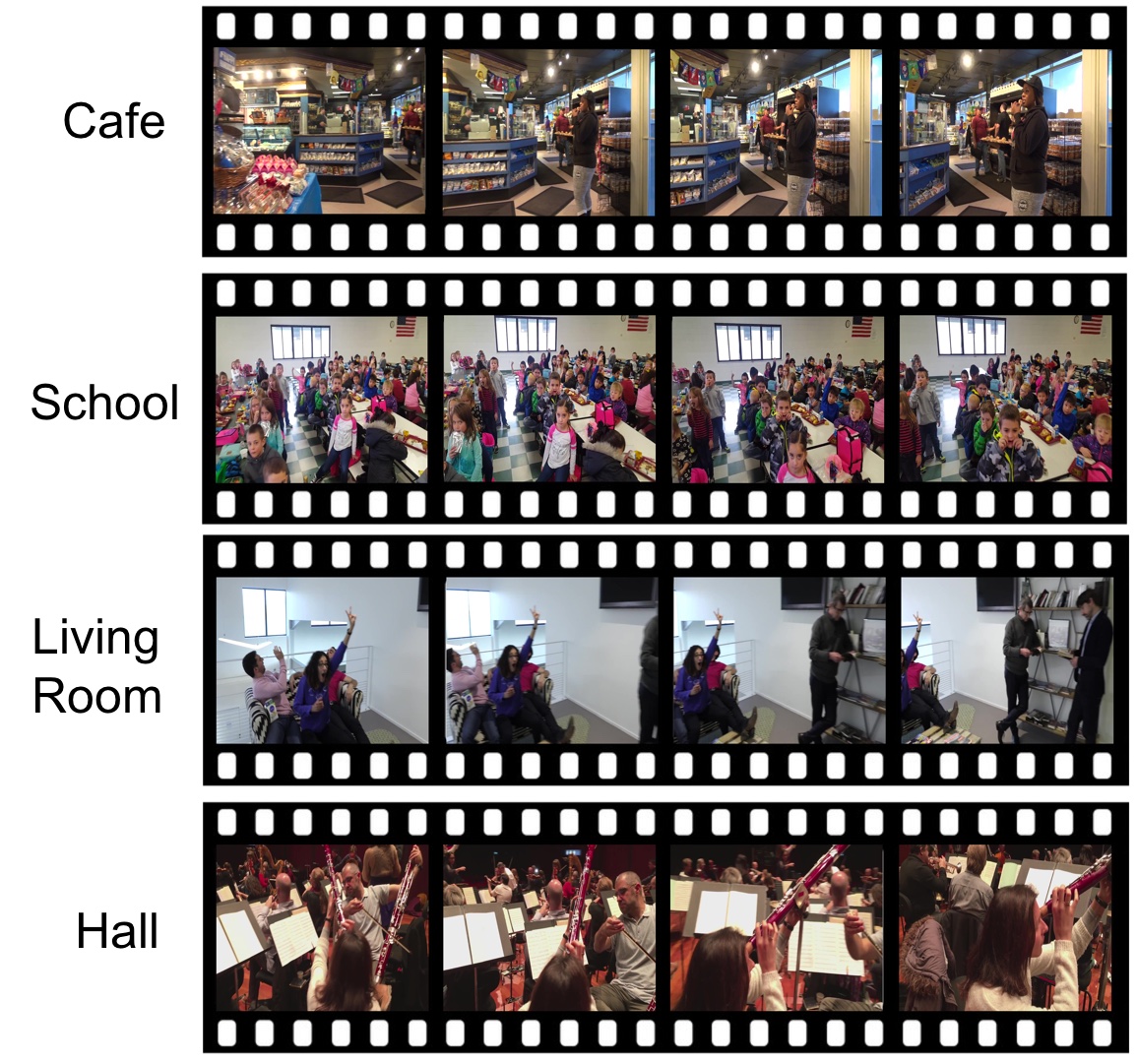}
\caption{Exemplary video clips from the Mannequin dataset, where
action-frozen people keep certain poses in diverse environments such 
as cafe, school, living room, hall, etc.} \label{fig:diversity}
\end{figure}

The main objective of our current research is to obtain high quality 3D
skeletons from multiple 2D poses obtained in each frame of action-frozen
people video.  Such a dataset was collected by focusing on a special
kind of Youtube video. That is, people imitate mannequins and freeze in
elaborate and natural poses in a scene while a hand-held camera tours
the scene to create the desired video. It is called the Mannequin
dataset. This type of video is suitable for human pose estimation since
it can provide diverse poses from people of different ages and genders
with different scales in a wide variety of outdoor scenes.  We propose a
multi-view matching (MVM) method to achieve this goal.  With a large
amount of in-the-wild video data labeled by 3D supervisions
automatically generated by MVM, we are able to train a neural network
that takes a single image as the input and generate the associated
multi-person 3D pose estimation as the output. 

Our approach has several advantages in multi-person 3D pose
estimation.  First, the scene does not have to be an indoor and/or lab
environment. Second, there are a considerable number of video clips
available online. If needed, it is possible to make more action-frozen
people video at a low cost. Third, we can get high quality 3D skeleton
information by exploiting the static scene assumption. To the best of
our knowledge, there is only one in-the-wild 3D human pose dataset,
called MPII-3DPW \cite{von2018recovering}. It relies on Inertial Measurement Unit
(IMU) sensors
attached to a few actors or actresses. As compared to MPII-3DPW, our
approach is more scalable. It contains more diverse contents in terms of
subjects and environments. Some exemplary video clips in the Mannequin
dataset are shown in Fig. \ref{fig:diversity}. With the help of such
video clips, we can obtain diverse 3D human poses in a wide range of
scenes in daily life. 

Estimating 3D skeletons from predicted 2D poses of multiple views has
been studied previously in several settings, e.g. the synchronized
multi-camera lab environment \cite{dong2019fast,joo2017panoptic}, the
synthesized sports video \cite{bridgeman2019multi}. However, existing
methods do not work well in the current setting (namely, video of
action-frozen people captured by a hand-held camera) for the following
reason. In the action-frozen video clip, each frame can be essentially
regarded as a single view of the scene. A typical 10-second video
sequence sampled at 25fps will yield 250 views of the scene from 250
viewing angles.  As observed in \cite{dong2019fast}, the main challenge
is to build the correspondence of predicted 2D poses in different
frames. The optimization method in \cite{dong2019fast} was developed to
minimize the cycle consistency loss in a multi-camera lab environment
with a small number of views (e.g. five cameras). The solution in
\cite{dong2019fast} is computationally intractable in our current case,
which has a larger number of views and more people in the scene. 

Besides, pose tracking methods like \cite{andriluka2018posetrack} can be a potential option when we need to determine which 2D poses in different frames corresponds to the same person. However, we argue that they are also not efficient enough in our case. The reason is that, when identifying the correspondence of multiple 2D poses, the 3D geometric constraints plays a more important role compared to the visual similarity clue that is usually what the pose tracking methods rely on, which is further confirmed by our empirical experiments. As it is not common for pose tracking methods to make assumption of static scenes, the 3D geometric constraints are not fully utilized. 

\begin{figure}[h]
\centering
\includegraphics[width=\textwidth]{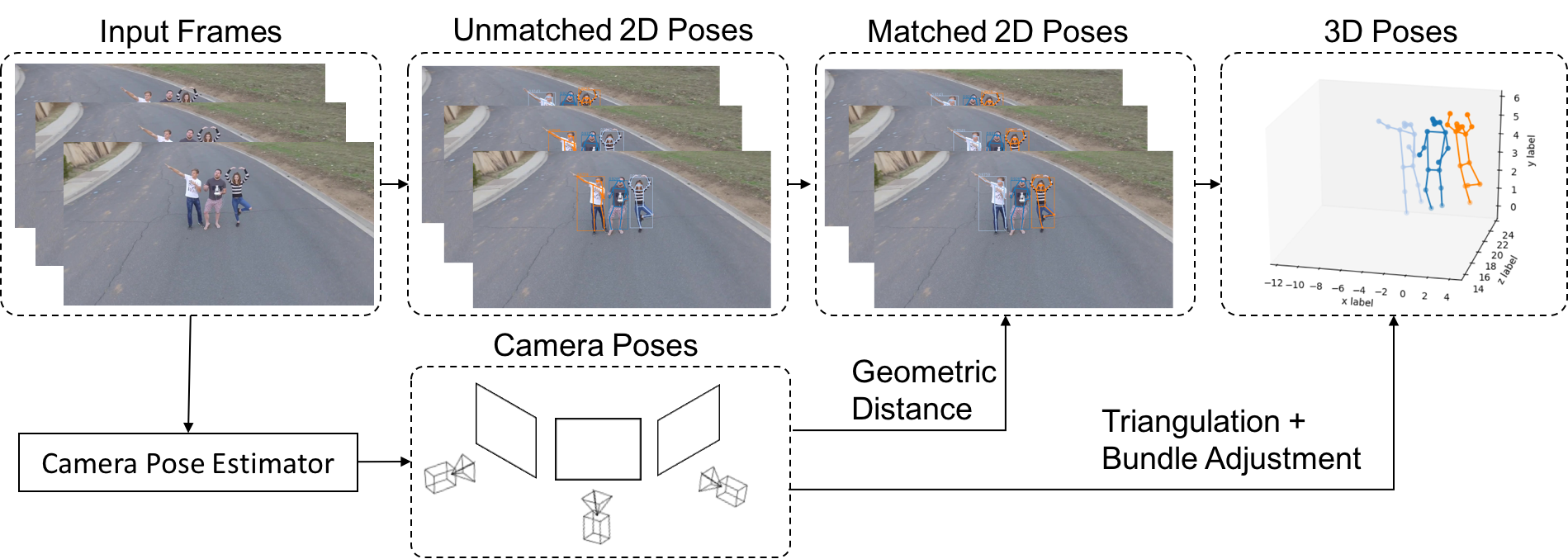}
\caption{Illustration of the generation of multi-person 3D poses
by the proposed MVM method.  The input contains a sequence of image
frames (shown in the first block), 2D pose estimation at each frame
(shown in the second block), building correspondence of 2D poses across
multiple frames (shown in the third block), recovering 3D poses from
matched 2D poses (shown in the fourth block).} \label{fig:framework}
\end{figure}

The pipeline of the proposed MVM method is shown in Fig.
\ref{fig:framework}. First, we use the OpenPifPaf method
\cite{kreiss2019pifpaf} to estimate multi-person 2D poses at each
frame.  Second, we find the correspondence of 2D poses across multiple
frames.  This is achieved by adopting an approximation to the
optimization objective in \cite{dong2019fast}, where both geometric
constraints and appearance similarities are taken into account.  Third,
we apply triangulation to groups of matched 2D poses and conduct bundle
adjustment to recover multi-person 3D poses.  We provide both
qualitative and quantative results to show the quality of the
multi-person 3D poses in Youtube video. 

In practice, we are interested in multi-person 3D pose estimation
from a single image.  To accomplish this goal, we train a monocular
multi-person 3D pose network, which is a modification of the network
proposed in \cite{kreiss2019pifpaf}, from multiple frames using the 3D
supervisions obtained by MVM.  To demonstrate the effectiveness of 3D
supervisions provided by the MVM method, we conduct experiments on the
3DPW and the MSCOCO datasets. Evaluation shows performance improvement
in the 3D human pose estimation accuracy for the 3DPW dataset and the 2D
human pose estimation accuracy for the MSCOCO dataset. 

There are four main contributions of our current research.
\begin{itemize}
\item We develop an efficient method, called the MVM method, that
levarages geometric constraints and appearance similarities existing in
video clips of static scenes.  for reliable 3D human poses estimation. 
\item We collect a large-scale Youtube video clips containing
action-frozen people with in-the-wild scenes and call it the Mannequin
dataset. It can be used as a training dataset for multi-person 3D
pose estimation. 
\item Although there is no groundtruth, we use the MVM method to generate
weak 3D supervisions for the Mannequin dataset so that the dataset can
be used to train a multi-person 3D neural network that is applied to a
single image. 
\item We use the Mannequin dataset as the training dataset and show the
effectiveness of the weak 3D supervisions provided by MVM through
extensive experimental results conducted for the 3DPW dataset and the 2D
MSCOCO dataset. 
\end{itemize}

The rest of this paper is organized as follows. Related work is reviewed
in Sec. \ref{sec:review}. The Mannequin dataset is introduced in Sec.
\ref{sec:dataset}. The MVM method is proposed in Sec.
\ref{sec:method-1}. The solution to the multi-person 3D pose estimation
problem from a single image is discussed in Sec. \ref{sec:method-2}.
Experimental results are shown in Sec.  \ref{sec:experiments}. Finally,
concluding remarks are given in Sec.  \ref{sec:conclusion}. 

\section{Review of Related Work}\label{sec:review}

Several previous research related to our current work is reviewed below.

\subsection{3D Human Pose Estimation} 

There are two common representations for 3D human poses: the
skeleton-based representation and the parametric-model-based
representation.  The skeleton-based representation serves as the minimal
representation of human poses. It finds applications such as in sports
motion analysis \cite{ohashi2020synergetic}, among many others. One
approach to skeleton-based 3D poses estimation is the use of the
2D-to-3D lifting mechanism.  Martinez {\em et al.}
\cite{martinez2017simple} proposed a method that uses a cascade of
fully-connected (FC) layers to infer 3D skeletons from 2D skeletons.
Recently, Ci {\em et al.} \cite{ci2019optimizing} adopted a modified
graph neural network to achieve the same goal. Yang {\em et al.}
\cite{yang20193d} proposed a data augmentation method to synthesize
virtual candidate poses, which improves performance consistently.

Besides using RGB images as the input, the skeleton-based representation
can also be applied to range images for 3D human pose estimation.  For
example, Marin-Jimenez {\em et al.} \cite{marin20183d} examined a
convolutional neural network (CNN) based method that estimates 3D human
poses from depth images directly.  Zhang {\em et al.} \cite{zhang20193d}
proposed a clustering based method with hybrid features by integrating
the geodesic distance and the superpixel-based mid-level representation. 

The parametric-body-model-based representation offers richer information
of the human body. Early work used the SCAPE model
\cite{anguelov2005scape} that fits a body model to annotated 2D
keypoints in the image. More recently, the SMPL model
\cite{loper2015smpl} becomes popular.  Kanazawa {\em et al.}
\cite{kanazawa2018end} proposed a method that fits the SMPL model by
minimizing the re-projection error with respect to the 2D keypoint
annotations while taking the human pose prior such as joint angles and
shape parameters into account.  Lassner {\em et al.} \cite{lassner2017unite}
included the 2D silhouette projection in the loss function for data with
segmentation mask annotations (e.g., the MSCOCO dataset).  

Although they are useful in several contexts, both representations
cannot be used in our multi-person 3D pose estimation framework
directly. For the former, it is not easy to generalize the skeleton
model to tackle with a varying number of people in an image. For the
latter, the parametric model is too heavy for multi-person 3D pose
estimation since it has too many parameters.  It is worthwhile to
mention that there exists work that targeted at obtaining 3D human poses
directly.  For example, Papandreou {\em et al.}
\cite{papandreou2017towards} showed that a coarse-to-fine volumetic
representation is better than the simpliest K-by-3 vector
representation, where $K$ is the number of keypoints of the skeleton.  A
similar conclusion was observed in \cite{sun2018integral}, which further
proposed an integral regression on 3D skeletons. 

\subsection{Multi-Person Pose Estimation} 

To solve the problem of multi-person pose estimation from a single
image, we can categorize current methods into two types: the top-down
approach \cite{papandreou2017towards,he2017mask,toshev2014deeppose,
wei2019view, ning2017knowledge,yu2017monocular,xu2020multi} and the bottom-up
approach \cite{kreiss2019pifpaf,cao2017realtime}. The top-down approach
runs a person detector and then estimates body joints in the detected
bounding box. The associated methods benefit from advances in person
detectors and a vast amount of labeled bounding boxes for people.  The
ability to leverage the labeled data turns the requirement of a person
detector into an advantage. Yet, when multiple bounding boxes overlap,
most single-person pose estimators do not work well. Unfortunetely,
there are many such cases in the Mannequin dataset of our interest so
that the top-down approach is not applicable. In contrast with the
top-down approach, the bottom-up approach does not rely on a person
detector.  Kreiss {\em et al.} \cite{kreiss2019pifpaf} first estimated
each body joint and, then, grouped them to form a unique pose by a
method called the Parts Asscociation Field (PAF). 

\subsection{Human Pose Tracking} 

A single-frame multi-person pose estimator cannot ensure consistency of
identity across frames. Depending on how multiple frames are utilized,
we categorize pose tracking methods into two types: offline methods and
online methods. To address the identity correspondence problem across
frames, Many offline methods \cite{insafutdinov2017arttrack,
iqbal2017posetrack, xiao2018simple, zhang2014robust} were proposed to
enforce temporal consistency of poses in video clips. Since it usually
demands the solution of some difficult-to-optimize formula in form of
spatio-temporal graphs, their solution speed is slow. 

For online methods, a common technique to handle the multi-person
identification problem across frames is to maintain temporal graphs in
neural networks \cite{doering2018joint,girdhar2018detect,
sedai2013gaussian}.  Rohit {\em et al.} \cite{girdhar2018detect}
proposed a 3D extension of Mask-RCNN, called person tubes, to connect
people across time. Yet, its tracking result is no better than the
simple baseline Hungarian Algorithm \cite{iqbal2017posetrack} even more
time and memory are needed to support the grouped convolution operations
in the implementation for a couple of frames. Joint Flow
\cite{dong2019fast} exploited the concept of the Temporal Flow Field to
connect keypoints across two frames. However, the flow-based
representation suffers from ambiguity when subjects moved slowly (or
being stationary). It demands special handling of such cases in the
tracking process. Apparently, this method is not applicable to the
Mannequin dataset that contains action-frozen people. 

\subsection{Design Choices of Proposed MVM}

In the proposed MVM method, we choose an approximating offline
optimization method to balance accuracy and computing efficiency.
Furthermore, we choose the bottom-up method since it is a better fit for
the context of our interest. Furthermore, we adopt the
``2D-skeleton-plus-relative-depth" representation for multi-person 3D
pose representation. The above design choices have several advantages.
First, unlike most top-down methods which are difficult to generalize to
scenes containing a varying number of people in an image, our method can
scale from the one-person case to the multi-person case elegently.
Second, both 2D skeletons and their depth information can be expressed
using local image coordinates projected to the camera matrix for each
corresponding frame independently. This greatly relieves the dependence
on camera's locations. As a result, our multi-person 3D pose estimation
network can be trained in the camera-location-independent setting. In
other words, our method can work in scenes with simple camera
caliberation. 

\section{Mannequin Dataset}\label{sec:dataset}

\begin{table}[h]
\centering
\begin{tabular}{ll} \hline \hline
        \# of & Count \\ \hline
        Sequences & 100 \\
        Sampled clips & 550 \\
        Frames & 47263 \\ \hline
        Unmatched 2D Poses & 525901 \\
        Matched 2D Poses & 176876 \\
        Generated 3D Poses & 2172 \\ \hline
        Ave. clips per sequence & 5.5 \\
        Ave. frames per clip & 85.9 \\
        Ave. 2D joints per triangulation & 41.3 \\ \hline \hline
\end{tabular}
\caption{Statistics of the Mannequin dataset.}\label{tab:stats}
\end{table}

\begin{figure}[h]
\centering
\begin{subfigure}[b]{0.49\textwidth}
\includegraphics[width=\textwidth]{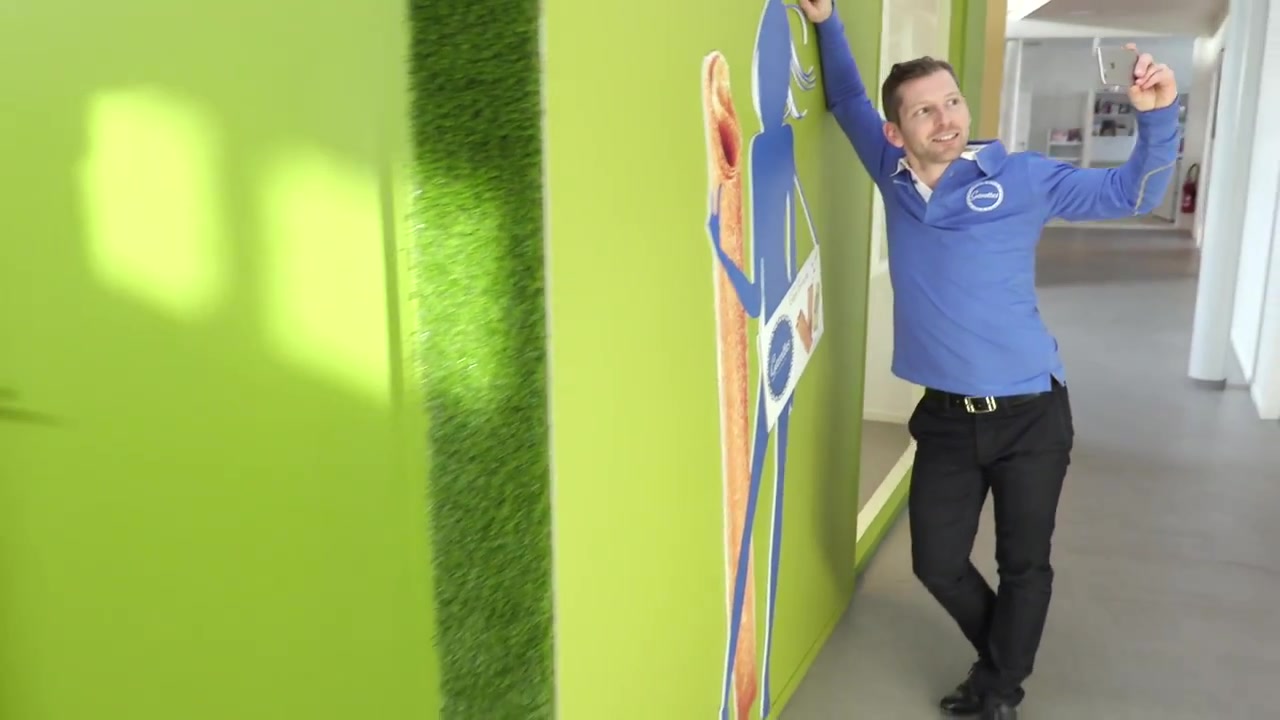}
\end{subfigure}
\begin{subfigure}[b]{0.35\textwidth}
\includegraphics[width=\textwidth]{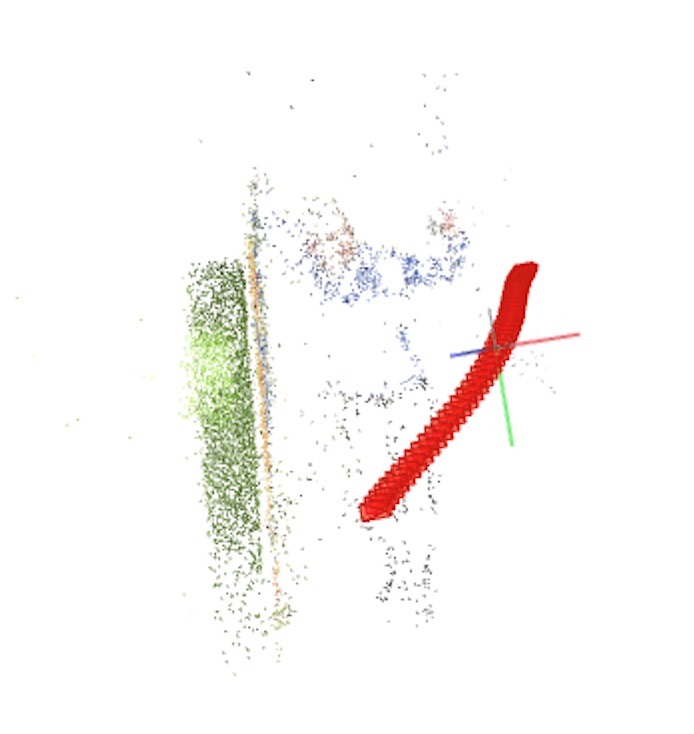}
\end{subfigure}
\begin{subfigure}[b]{0.49\textwidth}
\includegraphics[width=\textwidth]{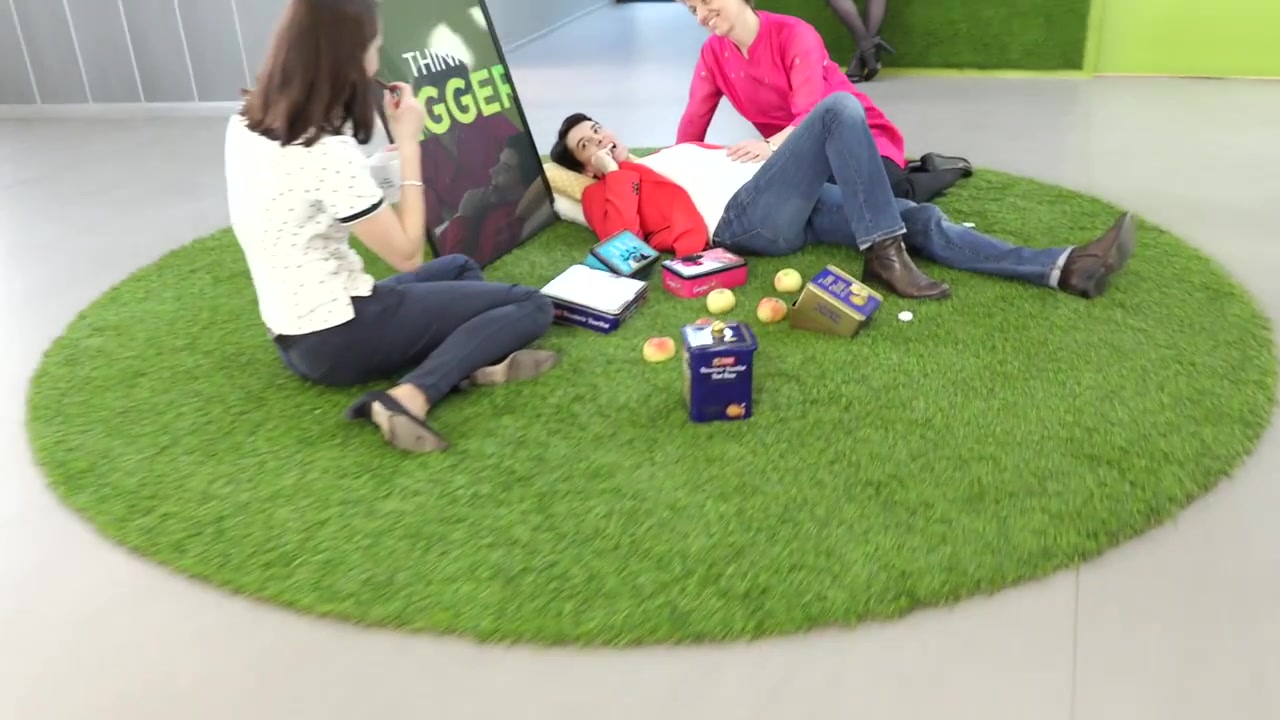}
\end{subfigure}
\begin{subfigure}[b]{0.35\textwidth}
\includegraphics[width=\textwidth]{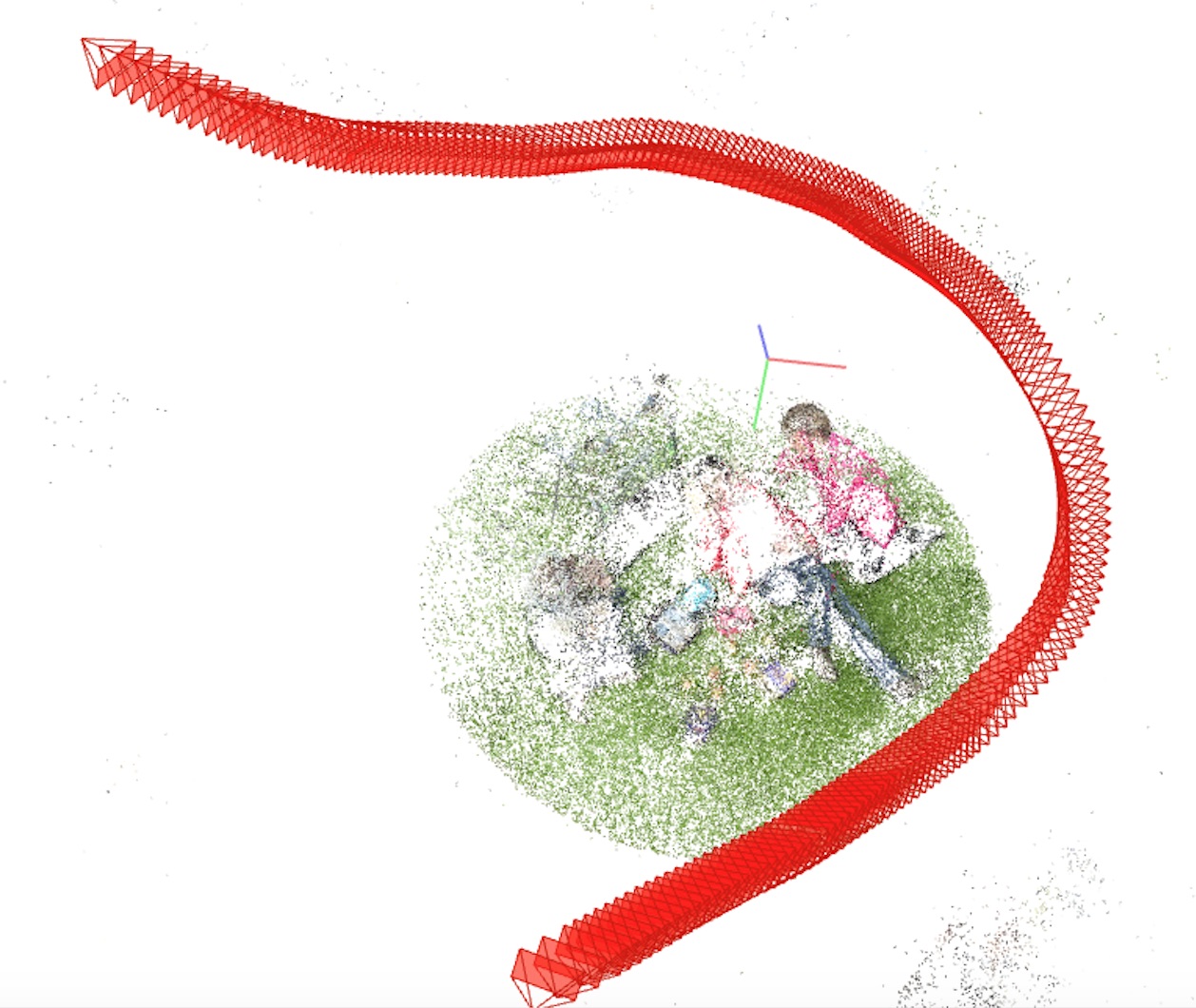}
\end{subfigure}
\caption{Visualization of estimated camera poses: exemplary image frames
in two video clips (left) and the corresponding estimated camera pose
trajectory curves in red (right).} \label{fig:colmap}
\end{figure}

The original Mannequin dataset \cite{li2019learning} has
around 2000 sequences which consist of around 170K annotated frames with
both camera poses and dense depthmaps available. However, some of the
video clips are no longer available.  We do not need the provided dense
depthmaps. To build our own dataset, we manually choose 100 sequences
and collect ten clips for each sequence, where each clip is less than 10
seconds. Our dataset consists of 47,263 frames in total. As compared
with the original Mannequin dataset, we only use the camera
intrinsics information. This is because our re-sampled frames have very
few overlaps with the annotated frames in the original sequences and we
need to re-compute the camera pose information using the COLMAP method
\cite{schonberger2016pixelwise} in sampled frames. Unlike the original
application of the Mannequin dataset in depth estimation which
requires geometric consistency of most pixels in a frame, we only demand
consistency of a few pixels corresponding to 2D keypoints. Thus, we have
a different criterion in selecting valid frames. 

The 100 sequences are chosen such that they contain a large amount of
body motion, pose and appearance variations. They also contain severe
body part occlusion and truncation. They are attributed to occlusions
with other people or objects and the fact that persons may disappear
partially or completely and re-appear again. The person scale also
varies across the video clip due to camera's movement. Thus, the number
of visible people and body parts varies across a video clip. 

\section{Multi-View Matching (MVM) Method}\label{sec:method-1}

The overall pipeline of the proposed 3D multi-view matching (MVM) method
is illustrated in \ref{fig:framework}. For each input clip, we first
estimate the camera poses for all frames. Then, a 2D pose estimator is
independently applied to those frames to get the initial 2D pose
predictions. Next, we run the matching algorithm to find the
correspondence of 2D poses across frames. After that, we use
triangulation on 2D poses that belong to the same person to get the
initial 3D skeleton. Finally, the 3D skeleton is further finetuned by
the bundle adjustment algorithm for better consistency of the estimated
3D pose. 

\subsection{Camera Pose Estimation} 

By following an approach similar to that adopted by
\cite{li2019learning} and \cite{zhou2018stereo}, we use ORB-SLAM2
\cite{mur2017orb} to identify trackable sequences in each video clip and
estimate the initial camera pose for each frame. In this stage, we set
the camera intrinsics the same as the one provided in the original
Mannequin dataset, and process a lower-resolution version of
the video clip for efficiency. Afterwards, we process each sequence at a
higher resolution using a visual SfM system
\cite{schonberger2016structure} to refine initial camera poses and
intrinsic parameters. This method extracts and matches features across
frames. Then, it conducts the global bundle adjustment optimization.
Our implementation is based on an open-sourced multi-view stereo (MVS)
system called COLMAP \cite{schonberger2016pixelwise}. Two camera pose
estimation examples are shown in Fig. \ref{fig:colmap}. We show an
exemplary image frame in the left and the estimated camera pose
trajectory curve of for the corresponding video clip in red in the
right. 

\subsection{2D Pose Estimation}\label{subsec:2dpose}

We use the OpenPifPaf method in \cite{kreiss2019pifpaf} as the 2D human
pose estimation baseline network. It is trained using results from the
keypoint detection task of the MSCOCO dataset. For fair comparison with
other similar work in \cite{kolotouros2019learning} and
\cite{kocabas2019vibe}, we adopt a ResNet-152 \cite{he2016identity}
backbone feature extractor and run the 2D pose estimator on each input
frame independently to get the 2D pose predictions $x_{ij}$ along with
confidence score $w_{ij} \in [0,1]$. 

\subsection{Matching}

The objective of this matching step is to determine a set of 2D poses
that belong to the same person. This can be mathematically stated as
follows. Suppose that there are $T$ frames in a video clip. We use
$x_{ij} \in [0,1]^{C\times2}$ to indicate the $j^{th}$ 2D pose in the
$i^{th}$ frame with $C$ joints for each 2D pose in the normalized image
coordinates.  For each person $k$, we would like to determine a group of
2D poses, denoted by
$$
G_k = \{ x_{i_1j_1}, x_{i_2j_2}, \cdots, x_{i_{M_k}j_{M_k}} \},
$$ 
that are associated with the same person and $M_k$ is the number of
frames this person shows up. 

\subsubsection{Affinity Matrix}

The matching criterion is based on the affinity function, denoted by
$A(x_u, x_v)$, of two 2D poses $x_u$ and $x_v$. The affinity
function takes two factors into account; namely, appearance 
similarity $S$ and geometric distance $D$. It is expressed as
\begin{equation} \label{eq:1}
A(x_u, x_v) = S(x_u, x_v) \times \frac{1}{1 + \exp{\big( \gamma D(x_u,
x_v) \big)}},
\end{equation}
where $\frac{1}{1+\exp{(\gamma D)}}$ converts a distance measure to 
a similarity measure and
\begin{equation} \label{eq:2}
D(x_u, x_v) = \frac{1}{2C} \sum_{c=1}^C  d(x_u^c, L_{uv}(x_v^c)) + 
d(x_v^c, L_{vu}(x_u^c))
\end{equation}
is a geometric distance measure between two poses. Furthermore, in Eq.
(\ref{eq:2}), $L_{uv}(x_v^c)$ indicates the epipolar line of the
$c^{th}$ joint of 2D pose $x$ from view $u$ to view $v$ and $d(.)$ is
the Euclidean distance between a point and a line. 

The appearance similarity, $S(x_u, x_v)$, in Eq. (\ref{eq:1}) is
calculated using the cosine similarity of the features extracted from
the last conv layer of the network described in Sec.
\ref{subsec:2dpose}. The geometric distance, $D(x_u, x_v)$, compute the
average distance between the epipolar lines of the 2D keypoints in one
frame and the corresponding 2D keypoints in the other frame. The overall
affinity is a product of the appearance affinity and the geometric
affinity as shown in Eq. (\ref{eq:1}). 

\subsubsection{Mutual Consistency Maximization}

Suppose that there are $N$ people in total in the entire set of input
frames. We use $y_{ij} \in [1..N]$ to indicate the associated person
index of pose $x_{ij}$. Our goal is to maximize the following objective
function:
\begin{maxi} 
{y}{\sum_{k=1}^N  \sum_{(i_1,j_1)\in G_k} 
\sum_{(i_2,j_2)\in G_k} w_{i_1j_1}w_{i_2j_2}A(x_{i_1j_1}, x_{i_2j_2}),}{}{}
\addConstraint{\{G_k\}}{\text{ is a partition for } \{x_{ij}\}.}{}{}
\label{eq:consistency}
\end{maxi}

In principle, Eq. (\ref{eq:consistency}) can be solved by finding the
partition of the affinity matrix with the spectral clustering method.
However, this classical method is too slow to be practical in our
current context.  To speed up this optimization process, we develop a
greedy algorithm that finds an approximation to the original objective.
The main idea is described below in words.  The algorithm maintains a
set of corresponding 2D poses. It begins with the 2D pose that has the
largest confidence from the pool of 2D poses, where the confidence score
is generated by the 2D pose network. At each time, we choose the 2D pose
with the highest affinity score from the pool, and repeat the process
until we cannot find any $x_{ij}$ that has an affinity score above
threshold $\tau$.  The pseudo codes of the proposed greedy 2D poses
matching algorithm are given in Algorithm \ref{alg:matching}. 

\begin{algorithm}[h]
\SetAlgoLined
\SetKwInOut{Input}{Input}
\SetKwInOut{Output}{Output}
\Input{2D Poses $x_{ij}$, Affinity Matrix $A$}
\Output{$G_k$ for $k \in [1..N]$}
 Initialize visited set $V = \emptyset$\;
 \For{$k \gets 1$ \KwTo $N$}{
  Find $x_{i_0j_0} \notin V$ with the largest confidence\; 
  $G_{k} \gets \{x_{i_0j_0}\}$\;
  Find $x_{ij}$ such that $\sum_{x_{pq} \in G_k} w_{ij}
  w_{pq}A(x_{ij}, x_{pq})$ is the largest and $i \neq p$ 
  for any $x_{pq} \in G_k$.\;
  \While{Confidence of $x_{ij}$ is above $\tau$}{
   $G_k \gets G_k \cup x_{ij}$\;
   Find $x_{ij} \notin V$ such that $\sum_{x_{pq} \in G_k} 
   w_{ij}w_{pq}A(x_{ij}, x_{pq})$ is the largest and $i \neq p$ 
   for any $x_{pq} \in G_k$.\;
   }
  $V \gets V \cup G_k$\;
 }
 \caption{Pseudo codes for greedy 2D poses matching.}
 \label{alg:matching}
\end{algorithm}

\subsection{Triangulation and Bundle Adjustment}

The reconstruction of 3D poses from a group of corresponding 2D poses is
a well-studied question in the 3D geometry literature. Here, we use the
direct linear transform (DLT) algorithm \cite{hartley2003multiple} to
estimate the 3D keypoints from multiple corresponding 2D keypoints.
Moreover, we use the RANSAC algorithm to eliminate outliers. The
triangulation is applied to different joints independently.
Triangulated 3D poses may suffer from small parallax angles, which
results in a very large reconstruction error. Some reconstructed poses
are even not possible for human beings. 

The common practice is to apply bundle adjustment on triangulated 3D
poses to get high quality estimates.  To implement this idea, we
minimize an error function that considers the reprojection error and the
human pose prior introduced in \cite{arnab2019exploiting} jointly. It is
formulated by a Gaussian mixture $(\mu_l, \Sigma_l)$ for $l=1, \cdots,
8$ on a dataset of diverse 3D poses \cite{joo2017panoptic} in form of
\begin{mini}
{X}{\sum_{x_{u}} E_R(X,x_u) + \lambda E_P(X),}{}{},
\addConstraint{X}{\in \mathbb{R}^{3 \times C},}{}{},
\label{eq:bundle}
\end{mini}
where
\begin{equation}
E_P(X) = -\log{\Bigg\{ \sum_{l=1}^8 \mathcal{N}(X|\mu_l,\Sigma_l) 
\Bigg\}}.
\end{equation}

After we obtain reconstructed 3D poses, we will project them back to
each frame to generate a 3D supervision that will be used in the
training of a 3D pose estimation network from a single image as
discussed in the next section.

\section{Multi-Person 3D Pose Estimation from Single Image}\label{sec:method-2}

\subsection{System Overview}

\begin{figure}[h]
\centering
\begin{subfigure}[b]{0.8\textwidth}
\includegraphics[width=\textwidth]{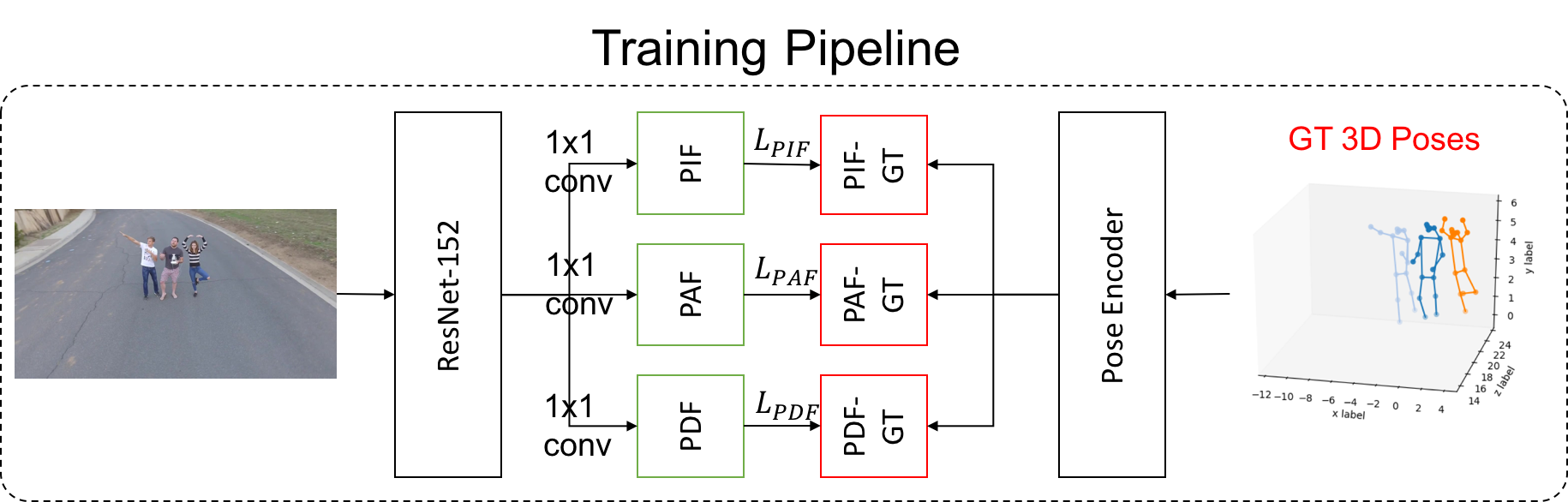}
\end{subfigure}
\begin{subfigure}[b]{0.8\textwidth}
\includegraphics[width=\textwidth]{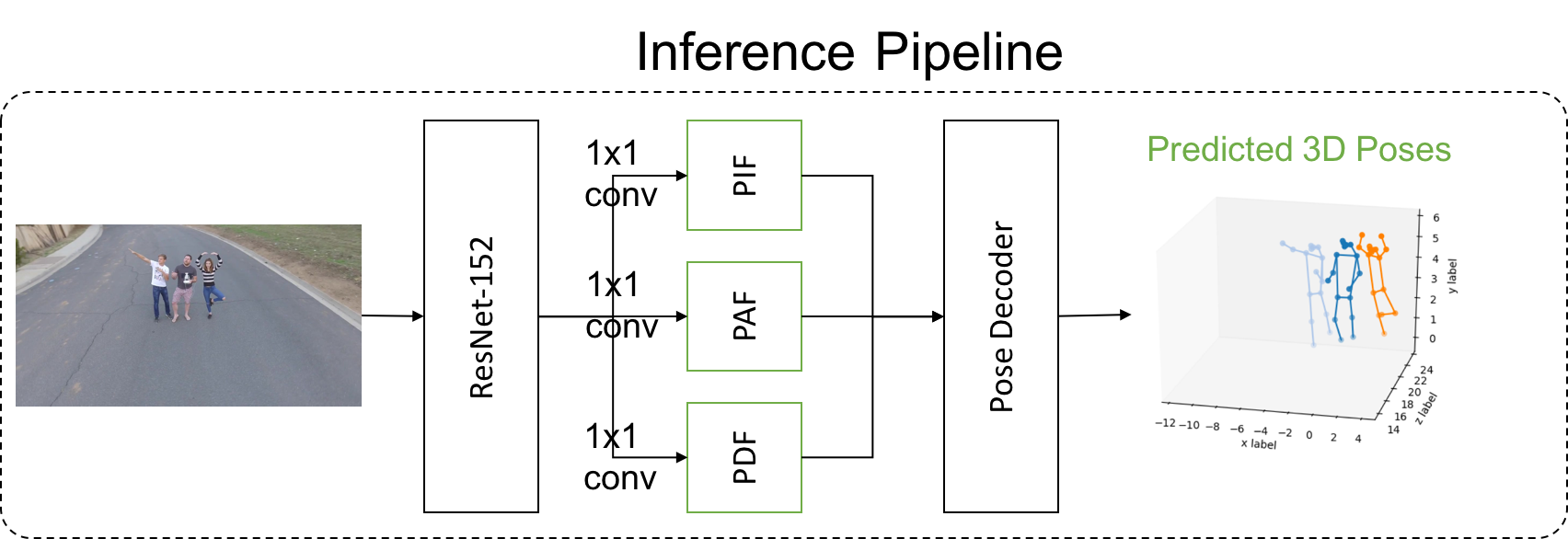}
\end{subfigure}
\caption{The proposed CNN for multi-person 3D pose estimation 
from a single image.}\label{fig:network}
\end{figure}

In this section, we study how to train a convolutional neural network
(CNN) to solve the problem of multi-person 3D pose estimation from a
single image. To address this problem, we represent estimated 3D poses
with two complementary components: 1) 2D poses and 2) keypoint depths.
Our CNN architecture is shown in Fig. \ref{fig:network}. It is a variant
of a state-of-the-art bottom-up human pose estimation network called
PifPaf \cite{kreiss2019pifpaf}. It has three key modules: 
\begin{itemize}
\item a backbone feature extractor (ResNet-152);
\item three prediction modules; namely, Parts Intensity Fields (PIF), 
Parts Association Fields (PAF), and Parts Depths Fields (PDF);
\item a 3D pose encoder/decoder.
\end{itemize}

It is worthwhile to mention that PIF and PAF are the same as
\cite{kreiss2019pifpaf} while PDF is new. Furthermore, most existing
multi-person 3D pose networks, e.g., \cite{arnab2019exploiting,
huang2019deepfuse}, take multiple frames as the input and perform some
sort of 2D-to-3D operations before producing the final 3D poses. In
contrast, our network can be trained in an end-to-end manner. 

\subsection{Key Components}

We will discuss the roles and implementations of PIF, PAF, PDF, and the
3D pose encoder and decoder here. 

PIF is used to describe where the 2D keypoints are so that the body
parts can be localized. Different joints are processed independently.
As far as the implementation is concerned, there are 5 components for
each location of joint $j$ in the output map. They are $c$, $p_x$, $p_y$,
$b$ and $\sigma$, where $c$ is the confidence score, $(p_x, p_y)$ denote
the coordinates of the point that is closest to joint $j$, $b$ is used
to characterize the adaptive regression loss for $(p_x, p_y)$, and
$\sigma$ stands for the standard deviation of the Gaussian component at
location $(p_x, p_y)$ when werecover the high-resolution confidence map
of joint $j$ from a low-resolution output map produced by the network. 

\begin{figure}[h]
\centering
\includegraphics[width=0.45\textwidth]{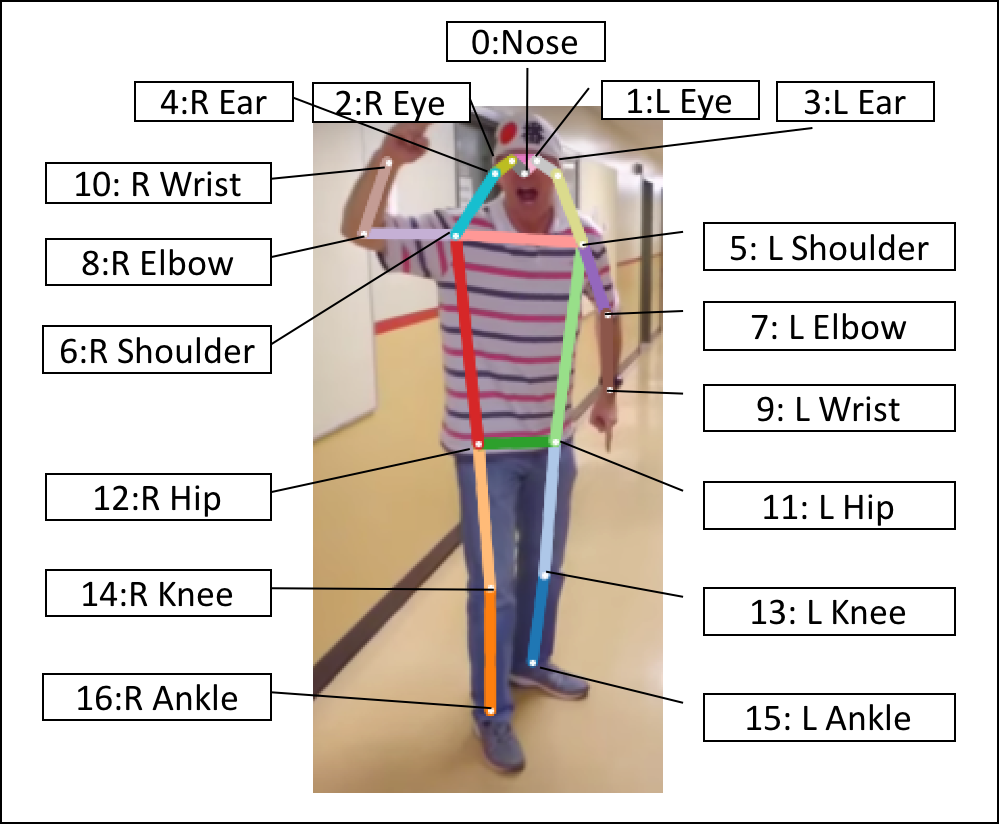}
\caption{The skeleton representation of human poses in our framework.}
\label{fig:skeleton}
\end{figure}

PAF is used to characterize the link location of a skeleton, where links
are used to connect joints detected by PIF. The joints and links are
associated together to form an instance of a skeleton. In our skeleton
representation as illustrated in Fig. \ref{fig:skeleton}, some PAFs
correspond to real bones while others do not. They are just some virtual
connections between joints, e.g. the connection between the left ear and
the left shoulder. Since both PIF and PAF are the same as those in
\cite{kreiss2019pifpaf}, we refer to the orginal paper for further
details. 

PDF is used to regress the relative depth of joints. We use the relative
depth because of inherent scale ambiguity of 3D poses recovered from
multiple 2D poses. Note that 3D poses used as the groundtruth are
expressed in the relative scale. Being similar to PIF, there are 2 PDF
components for each location $(i, j)$. They are $d_{ij}$ and
$\gamma_{ij}$.  The former represents the relative depth at locaiton
$(i, j)$ while the latter denotes the radius of the Gaussian component
centered at the location of the closest joint, denoted by $(p_x^{ij},
p_y^{ij})$.  The actual value of the high resolution depth map at
location $(x, y)$ satisfies the following:
\begin{equation}
D_{xy} = \sum_{i,j} d_{ij} \mathcal{N}(x,y|(p_x^{ij}, p_y^{ij}),\gamma_{ij}).
\end{equation}

The 3D pose encoder is used to encode three parts fields to the 3D poses
while the 3D pose decoder is used to decode the 3D poses to three part
fields. As to the 2D poses plus the depth representation of 3D poses, we
encoder/decode the 2D pose part in the same way as that is given in
\cite{kreiss2019pifpaf}. 

\subsection{Loss Function}

Since the 3D pose encoder is not differentiable, we need to compute the
loss between encoded groundtruth parts fields and the predicted parts
fields.  The overall loss function is given below:
\begin{equation}\label{eq:loss_all}
L = \lambda_1 L_{PIF} + \lambda_2 L_{PAF} + \lambda_3 L_{PDF}.
\end{equation}

Because our camera poses are estimated through the SfM-based method,
scale ambiguity is inevitable in our case. To solve the ambiguity
problem, we use the following relative depths loss in our network:
\begin{equation}\label{eq:loss}
\begin{split}
L_{PDF}(d_1, d_2) &= Var[\log{\frac{d_1}{d_2}}] \\ &= 
\frac{1}{N}\log{\big( \frac{d_1}{d_2} \big) }^2 - \frac{1}{N^2} 
\big(\sum{\log{ \frac{d_1}{d_2}}}\big)^2.
\end{split}
\end{equation}
Therefore, our estimated keypoint depths will try to fit the relative
scale of the depth label. It means that the estimated depths and the
depth labels are the same if the ratio of these two depths at all
locations are consistent. 

\subsection{Length Scale Calibration}

Although the predicted depths are relative, we can still make use of the
prior information hidden in human poses to have a rough estimation of
the missing scale, $s$. We need to find $s$ such that the squared
difference between the scaled bone length $sl_i$ and the average bone
length $\bar{l_i}$ is minimized as given below. 
\begin{mini} 
{s}{\sum_{i} \frac{1}{2}(sl_i - \bar{l_i})^2 }{}{}
\label{eq:scale}
\end{mini}
It is not difficult to show that Eq. (\ref{eq:scale}) has the optimal solution 
at
\begin{equation}
s^* = \frac{\sum_i \bar{l_i}l_i}{\sum_i l_i^2}.
\end{equation}

\subsection{Network Training}

Our proposed CNN for multi-person 3D pose estimation from a single image
is shown in Fig.  \ref{fig:network}. It can be trained by data with
imcomplete labels, e.g. images with 2D supervisions only, human pose
labels with some missing joints. This training flexibility comes from
the pose encoder that can generate independent maps for PIF, PAF and PDF
separately. If some fields are missing in the ground truth labels, we
can simply set the weights of these fields to zero while keeping labels
of other fields in the gradient computation process.  When training
images have 3D supervisions, we project the 3D skeleton information back
to the view of each frame so that our network can learn camera-independent 
multi-person 3D poses. 

Before we train the proposed CNN with the Mannequin dataset, we first
train it with 64,115 images in the 2017 COCO training set that have 2D
pose annotations. In this pre-training process, the PDF branch is not
affected since the COCO dataset does not have the depth information.
The pre-training of 2D pose related fields can improve the robustness of
the ultimate 3D pose estimation network significantly. 

In the network training with the Mannequin dataset, we apply the same
data augmentation as that is done in \cite{kreiss2019pifpaf}. To create
uniform batches, we crop images into squares whose side is around
95-100\% of the short edge of the image with a randomly selected
location. The large crops are done to preserve the size of training
images as much as possible.  Training images and annotations are
horizontally flipped randomly. We use the SGD optimizer with a learning
rate of 0.001, momentum of 0.95, a batch size of 8 and no weight decay. We
employ model averaging to extract stable models for validation. At each
optimization step, we update an exponentially weighted version of the
model parameters. The decay constant is 0.001. 

\section{Experiments}\label{sec:experiments}

We first conduct experiments to verify the validity of the proposed MVM
method quantitatively with the Mannequin dataset.  Then, we do
performance benchmarking with two public datasets; namely, 3DPW and
MSCOCO. The purpose is to show how additional training based on the 3D
supervision offered by the proposed MVM method as well as the Mannequin
dataset helps improve the performance of the multi-person 3D pose
network. 

\subsection{MVM Evaluation with Mannequin Dataset}

\begin{figure}[h]
\centering
\begin{subfigure}[b]{0.32\textwidth}
\includegraphics[width=\textwidth]{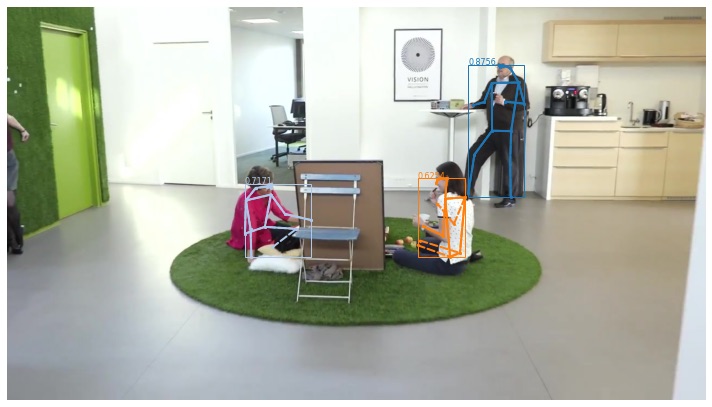}
\end{subfigure}
\begin{subfigure}[b]{0.33\textwidth}
\includegraphics[width=\textwidth]{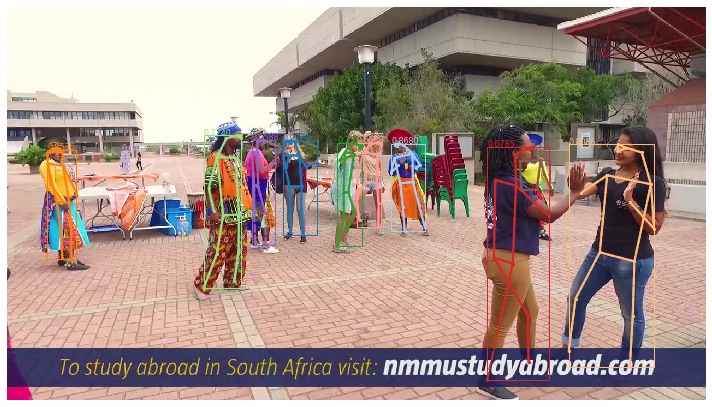}
\end{subfigure}
\begin{subfigure}[b]{0.33\textwidth}
\includegraphics[width=\textwidth]{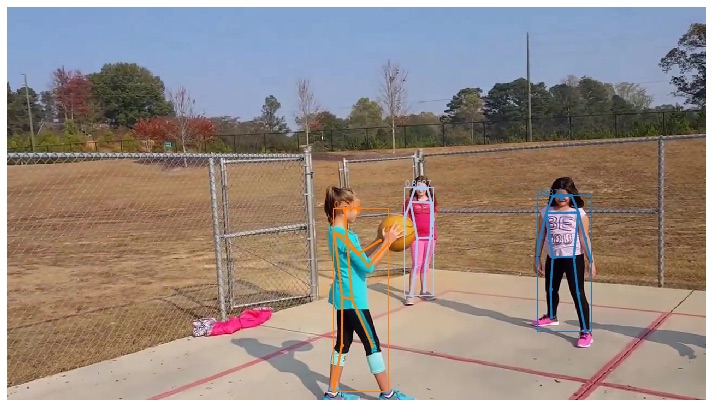}
\end{subfigure}
\begin{subfigure}[b]{0.32\textwidth}
\includegraphics[width=\textwidth]{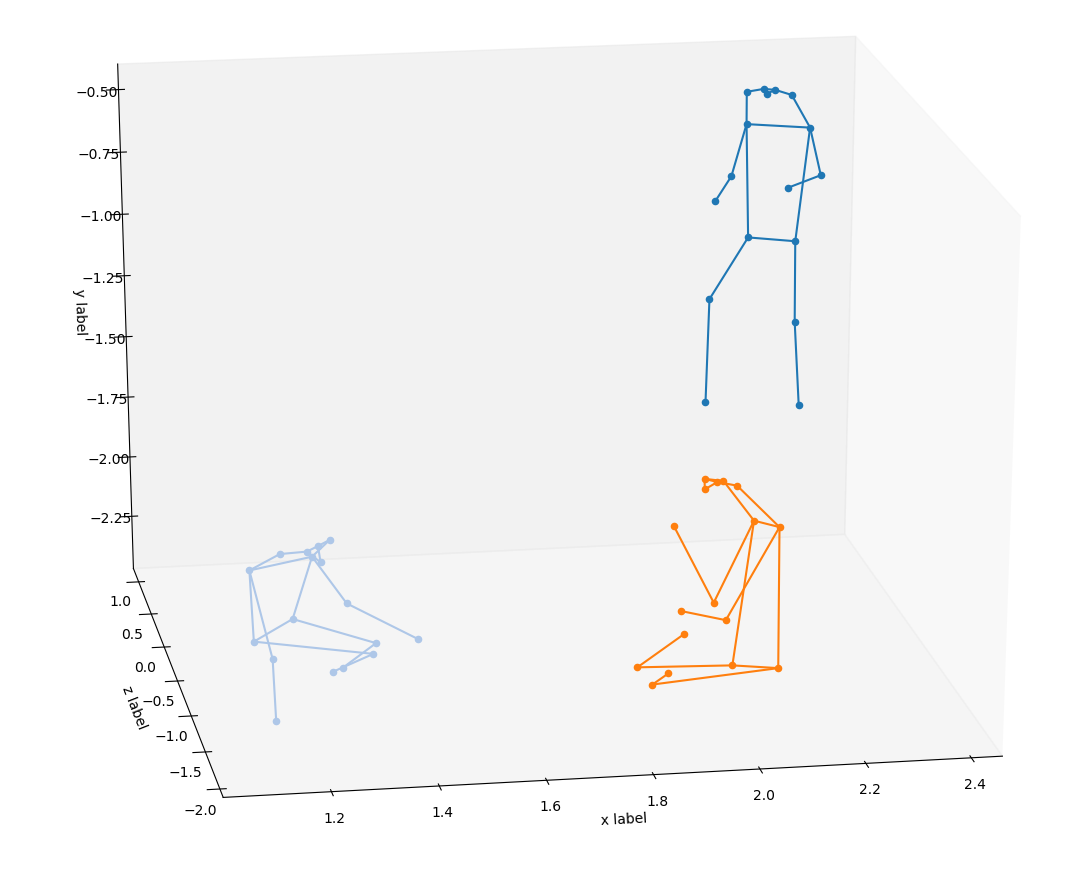}
\end{subfigure}
\begin{subfigure}[b]{0.33\textwidth}
\includegraphics[width=\textwidth]{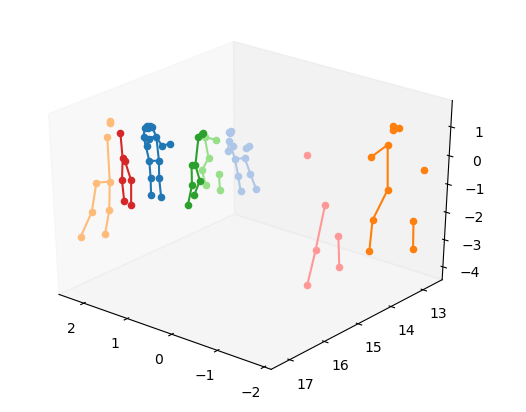}
\end{subfigure}
\begin{subfigure}[b]{0.33\textwidth}
\includegraphics[width=\textwidth]{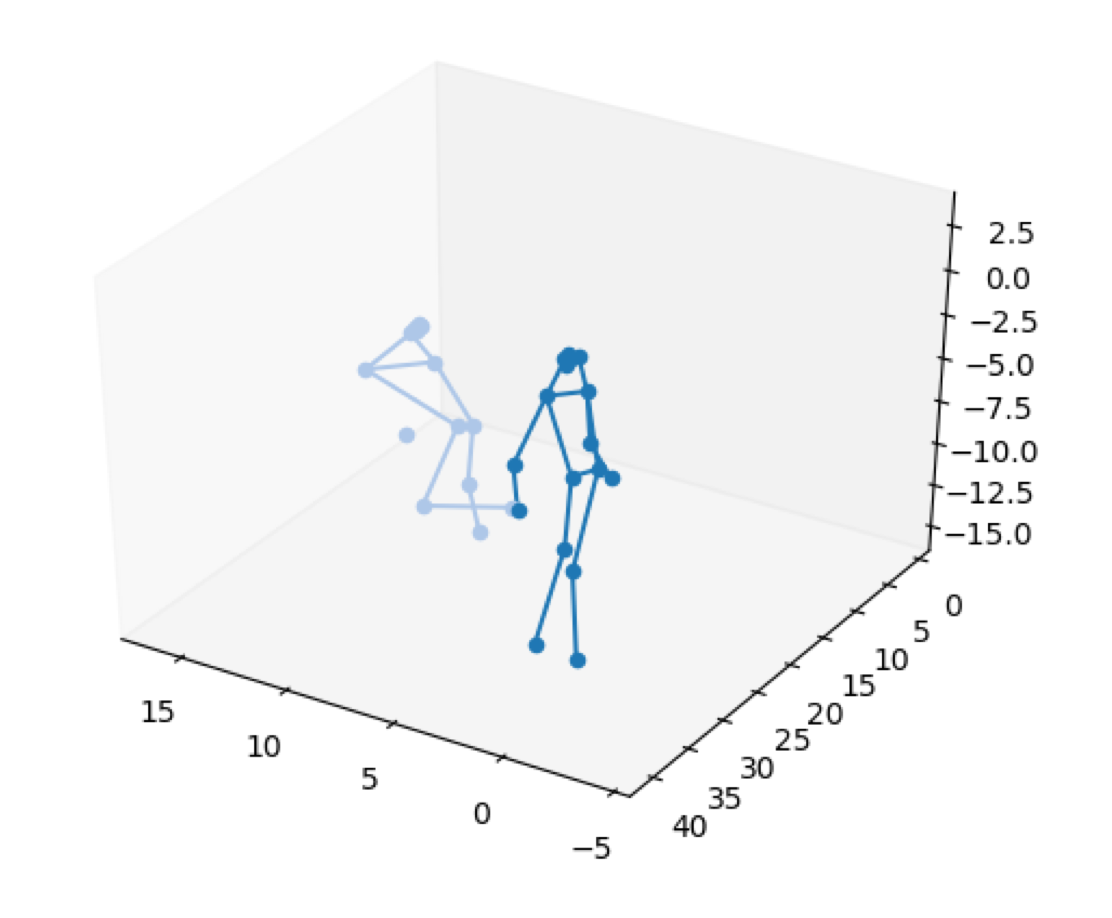}
\end{subfigure}
\caption{Visualization of the Mannequin dataset results: (1st row)
original images with 2D poses, (2nd row) generated 3D poses.  The first
column indicates a successful estimation of 3D poses in the scene. Due
to the low confidence score (e.g., the orange girl in the third column)
or a small parallax angle which can cause the failure of 3D
reconstruction (e.g., some part of the joints associated with the
orange-and-pink person in the second column), the MVM method
filters out such instances in a scene.} \label{fig:eval}
\end{figure}

We will demonstrate the effectiveness of the proposed MVM method using
the Mannquin challenge dataset in both qualitative and quantitative
ways.  For the qualitative performance of generated 3D poses, some
predicted examples are shown in Fig. \ref{fig:eval}.  There are two
people sitting on the carpet with severe occluded poses in the first
column of this figure.  We see from this example the capability of the
proposed MVM method in recovering complicated poses. In some scenarios,
we may have a low confidence score of a predicted human pose (e.g. the
orange girl in the third column) or a small parallax angle to result in
the failure of 3D reconstruction, (e.g., some part of the joints
associated with the orange-and-pink person in the second column).  The
MVM method tends to filter out such instances. 

For quantitative evalutions, since the Mannequin dataset does not
have any labels, we use the following metric to quantify the
validity of generated 3D poses:
the reprojection error $E_R$ of estimated 3D poses with their corresponding 2D poses. If the average reprojection error is low enough, it is reasonable to trust quality of the generated 3D poses. 

\begin{table}[h]
\hspace*{-1.5cm}
\centering
\begin{tabular}{l|ccc|cc} \hline \hline
Matching Algorithm & $E_R$ (px) $\downarrow$ & Outliers $\downarrow$ 
& $|G_k|$ $\uparrow$ & $E_R$ (px) w/o RANSAC $\downarrow$ 
& $|G_k|$ w/o RANSAC \\ \hline \hline
Baseline (Hungarian + MSE) & 33.1 & 12.1 & 6.5 & N/A & 18.7 \\
Arnab et al. \cite{arnab2019exploiting} (Shortest Path w/ MSE) & 25.7 
& \textbf{3.3} & 14.7 & 40.9 & 18.1 \\
MVM & \textbf{14.6} & 8.5 & \textbf{41.2} & \textbf{28.8} & 49.7\\ \hline
*Dong et al. \cite{dong2019fast} (Optimization + Our Similarity) 
& \textbf{9.6} & \textbf{2.9} & 10.5 & \textbf{22.3} & 13.5 \\
*MVM & 10.4 & 5.1 & \textbf{10.8} & 25.1 & 15.9 \\ \hline \hline
\end{tabular}
\caption{Performance comparison between several matching algorithms,
where * means only evaluated on clips with less than 50 frames.
Otherwise, it would take too much time to run Dong's algorithm. Smaller reconstruction error $E_R$ implies more consistent 3D poses. A larger number of corresponding 2D keypoints $|G_k|$ as well as a smaller number of outliers means a more robust 3D reconstruction process and thus potentially better 3D poses.}
\label{tab:matching}
\end{table}


We compare the MVM method with three other methods in Table
\ref{tab:matching}. They are:
\begin{enumerate} 
\item a baseline that uses sequential matching;
\item a shortest path matching method \cite{arnab2019exploiting};
\item the state-of-the-art optimization-based matching algorithm \cite{dong2019fast}.  
\end{enumerate} 
The triangulation is computed joint-wise.  We also compare the effect of
RANSAC in eliminating outliers. The quantity, $|G_k|$, in the table
means the number of the 2D keypoints chosen for triangulation. 

As shown in Table \ref{tab:matching}, the proposed MVM method achieves
comparable performance with the state-of-the-art optimization method
\cite{dong2019fast} on short clips at a much lower computational cost.
These video clips are short enough for the optimization method to
converge within a reasonable amount of time (e.g. an hour). For longer
video clips, our MVM method outperforms both the baseline method and
the shortest path matching algorithm by a large margin. 

\begin{table}[htb]
\centering
\begin{tabular}{l|cc} \hline \hline
& Ave. $E_R$ & Ave. $|G_k|$ \\ \hline \hline
Shortest Path w/ MSE & 25.7 & 14.7 \\
Shortest Path w/ Geo Dist & 20.5 & \textbf{22.6} \\
Shortest Path w/ Geo Dist + Appearance & \textbf{19.6} & 19.8 \\ \hline
MVM w/ MSE & N/A & 15.8 \\
MVM w/ Geo Dist & 16.8 & \textbf{45.5} \\
MVM w/ Geo Dist + Appearance (ours) & \textbf{14.6} & 41.2 \\ \hline \hline
\end{tabular}
\caption{Comparison among different affinity matrices.}
\label{tab:sim}
\end{table}

Furthermore, we show how different similarity metrics affect the quality
of generated 3D poses in Table \ref{tab:sim}. It is clear from the table
the proposed geometric distance contributes the most performance gain in
terms of the average reprojection error. As to the appearance
similarity, although it reduces the total number of matched 2D poses
slightly, we can obtain the best overall reprojection error by combining
the geometric distance and the appearance similarity since their
integration introduces more constraints to matched poses. 

\subsection{3DPW Evaluation}

Very few multi-person 3D human pose datasets are available to the
public.  One recently released dataset, called the MPII-3DPW
\cite{von2018recovering}, contains 60 clips. It contains outdoor video
clips captured by a mobile phone with 17 IMUs attached to the subjects. The IMU data allow people to accurately
compute 3D poses and use them as the ground truth.  The test set
consists of 24 video clips. We use the 14 keypoints that are common in
both MSCOCO and SMPL skeletons. The same setting was also used in
\cite{arnab2019exploiting}. We evaluate on those frames that have enough
visible keypoints for 3D pose estimation and ignore subjects that have
less than seven 2D visible keypoints. We compute the Procrustes Aligned
Mean Per Joint Position Error (PA-MPJPE) \cite{arnab2019exploiting}
independently for each pose and, then, average errors for each tracked
person in each video clip. This process implies that we count video
clips with two people twice, etc. Finally, we average over the entire
dataset. 

\begin{table}[h]
\centering
\begin{tabular}{l|cc}\hline\hline
single frame & PA-MPJPE (mm) $\downarrow$  \\ \hline
Popa et al. \cite{popa2017deep} & 108.2 \\
SMPLify, Bogo et al. \cite{bogo2016keep} & 108.1 \\
Train w/ Our network & 82.3 \\
Train + Mannequin dataset w/ our network & \textbf{78.8} \\ \hline
Train w/ HMR \cite{kanazawa2018end} & 81.3  \\
Train + Mannequin dataset w/ HMR \cite{kanazawa2018end} & \textbf{78.2} \\ \hline \hline
\end{tabular}
\caption{Evaluation of estimated 3D poses for the MPII-3DPW dataset.}\label{tab:3dpw}
\end{table}

Table \ref{tab:3dpw} shows how incorporating additional data from our
Mannequin dataset improves results on 3DPW testing set.  Training our
proposed network with our data can improve the PA-MPJPE by 3.5mm. Since
3DPW dataset has annotations for the full parametric body model, SMPL
model \cite{loper2015smpl}, it is expected to have better performance if
the network can leverage the such information. Thus, we compare the
performance of HMR model trained in 3DPW dataset and the performance
trained in 3DPW and Mannequin dataset. The PA-MPJPE was improved by 3.1mm.

\subsection{MSCOCO Evaluation}

Besides facilitating 3D pose estimation, we are able to improve the
accuracy of 2D pose estimation by training a learning system using our
Mannequin dataset.  Table \ref{tab:evalution2} provides quantitative
results for the COCO dataset. The evaluation metrics in COCO dataset  are based on the mean average precision (AP) over 10 object keypoint similarity (OKS) thresholds as the main competition metric \cite{lin2014microsoft}. The OKS score can measure the similarity between two 2D poses, which essentially serves as a similar role like what IoU does in object
detection or segmentation. Our dataset can improve the baseline
performance by 1.9 in mAP. The training benefits from the additional data in our dataset which has a lot of weird poses.  Apparently, it is more likely for people to make strange and
challenging poses in the shooting of a video clip to be uploaded to the
Youtube. As a result, our dataset helps recover difficult cases. 

\begin{table}[h]
\centering
\begin{tabular}{l|cc} \hline \hline
& mAP@OKS $\uparrow$  & AP@OKS=0.5 $\uparrow$ \\ \hline
Train & 64.6 & 85.9 \\
Train + Mannequin dataset & \textbf{66.5} & \textbf{87.1} \\ \hline \hline
\end{tabular}
\caption{Evaluation of estimated 2D poses in the COCO validation set.}\label{tab:evalution2}
\end{table}


These experiments show how one can effectively use the Mannequin dataset
to improve the per-frame 2D pose model in multiple datasets.  Besides, in Table
\ref{tab:2d-consistency}, we compute $C_{var}$ to empirically compare the consistency of the initial 2D pose estimator. As is defined in Equation \ref{eq:var}, $C_{var}$ stands for the variance of pairwise triangulated 3D keypoints of 2D pose predictions,

\begin{equation} \label{eq:var}
    C_{var}(\{x_u^c\}) = \frac{1}{M(M-1)/2} \sum (F_{tri}(x_u^c, x_v^c) - \mu_X^c)^2,
\end{equation}
where $\tau$ is the threshold to eliminiate outliers.  As shown
in the table, OpenPifPaf can produce a more consistent set of 2D poses.
This is the reason we used OpenPifPaf as the default 2D pose estimator. 

\begin{table}[h] \centering
\begin{tabular}{l|cc}\hline\hline
2D pose network & $C_{var}@\tau=0.5$ & $C_{var}@\tau=0.9$ \\ \hline \hline
Mask-RCNN \cite{he2017mask} & 581.8 & 233.6 \\
OpenPifPaf \cite{kreiss2019pifpaf} & \textbf{351.0} & \textbf{185.7} \\ \hline \hline
\end{tabular}
\caption{Comparison between two 2D pose estimation networks, where we
report the variance, $C_{var}$, of the pairwise triangulated 3D poses,
which is used to measure consistency of predicted 2D poses, and $\tau$
is the threshold to eliminiate outliers.} \label{tab:2d-consistency}
\end{table}

\section{Conclusion and Future Work}\label{sec:conclusion}

An MVM method was proposed to generate reliable 3D supervisions from
unlabeled action-frozen video in the Mannequin dataset so as to improve
the estimation accuracy of multi-person 2D and 3D poses. The MVM method
attempts to match 2D human poses estimated across multiple frames. The
key to efficient matching lies in taking the geometric constraint (of
static scenes in the Mannequin dataset) and appearance similarity into
account jointly.  Afterwards, through the triangulation of a group of
matched 2D poses, optimization by considering the human pose prior and
re-projection errors jointly and bundle adjustment, we can obtain
reliable 3D supervisions.  These 3D supervisions are used to train a
multi-person 3D pose network.  It was demonstrated by experimental
results that both 3D pose estimation accuracy (tested for the 3DPW
dataset) and 2D pose estimation accuracy (tested for the MSCOCO dataset)
can be improved. 

There are several research directions worth further exploration. They are
elaborated below. 
\begin{itemize}
\item Acquiring more information from videos of action-frozen people
such as 3D surfaces, textures, etc. In the static scene setting, it is
feasible to obtain reliable estimates of various scene information with
no manual labels. 
\item Utilizing weak 3D supervisions obtained from the Mannequin
dataset.  Although the 3D poses generated by MVM method are accurate, we
may have an incomplete number of joints. This is caused by insufficient
parallax (viewing angle). To give an example, the right ear might be
missing since it is not seen in input frames. One might ``inpaint''
incomplete pose supervisions to obtain a pseudo groundtruth. 
\item Utilizing 3D point cloud information when running COLMAP.  COLMAP
can be used to estimate camera poses and the 3D point cloud information.
Currently, we only use the camera pose information.  The 3D point cloud
set may provide some extra useful information for better 3D pose
estimation. 
\end{itemize}

\bibliographystyle{unsrt}  
\bibliography{references}  


\end{document}